\documentclass[11pt,a4paper]{article}

\usepackage[utf8]{inputenc}
\usepackage{amsmath, amssymb, amsfonts}
\usepackage{graphicx}
\usepackage{booktabs}
\usepackage{multirow}
\usepackage{color}
\usepackage{hyperref}
\usepackage{cite}
\usepackage{algorithm}
\usepackage{algorithmic}
\usepackage{subfigure}
\usepackage{float}
\usepackage{bm}
\usepackage{ifpdf}

\usepackage[margin=1in]{geometry}

\title{Mahalanobis-Based Multi-Head Attention for Complex State Propagation}
\author{Xiaohe Li\thanks{GuangDong Police College} }
\date{\today}

\begin{document}

\maketitle

\begin{abstract}
In our previous work, we demonstrated that Complex State Propagator (CSP) alone suffices for deterministic state tracking tasks such as parity checking and parenthesis matching. However, CSP relies on input-dependent rotations in complex space, which inevitably introduces \textbf{synthetic distance distortions} across layers—especially when processing long sequences with nested structures. 

In this paper, we propose \textbf{Mahalanobis-Based Multi-Head Attention} (MHA-CSP), a novel attention mechanism that replaces the standard dot-product with a \textbf{Mahalanobis distance-based RBF kernel}, which effectively computes attention in an infinite-dimensional feature space without increasing the parameter count. Crucially, the positive definiteness of the Mahalanobis distance enables a \textbf{direct construction of Tree Attention}: attention scores are built directly from accumulated distances, with a LogSumExp correction that rectifies the raw distance by subtracting the log-sum of edge exponentials. Moreover, the multi-head Mahalanobis distance matrices are themselves repurposed to construct an \textbf{attention meshing mechanism}, enabling cross-head kernel collaboration that simultaneously boosts accuracy and training efficiency. Extensive experiments demonstrate that MHA-CSP, with only 119K parameters and \textbf{teacher forcing applied exclusively at the final hidden state}, consistently outperforms Transformer and GCN baselines trained from scratch under identical conditions on long-sequence state tracking tasks. While these baselines rely on dense attention or graph propagation, MHA-CSP achieves robust structured reasoning via synthetic distance rectification—powered by Mahalanobis-based attention—and efficient information bypass inherited from the CSP backbone. 

This result highlights the effectiveness of complex-valued state propagation with collaborative multi-head rectification in capturing symbolic structures, establishing a new efficiency-performance trade-off for structured reasoning. Our code is available at \url{https://github.com/hilhert/CSP-MHD}.

\end{abstract}

\section{Introduction}
Since its introduction, the Transformer architecture \cite{vaswani2017attention} has garnered significant attention and has become a de facto standard in sequence modeling, underpinning state-of-the-art systems across natural language processing, computer vision, and beyond. Its core innovation—the attention mechanism—enables models to dynamically weigh relationships between tokens, yielding remarkable flexibility and performance.

However, recent studies have revealed several intriguing phenomena regarding the internal workings of trained Transformers. Empirical observations indicate that the query (Q) and key (K) projection matrices often exhibit homogeneity \cite{bhojanapalli2020low}, and well-trained models frequently operate in a low-rank regime \cite{chen2021low, vonoswald2025weight}, suggesting that much of the parameter budget is underutilized. Concurrently, researchers have pointed out that standard Transformers struggle with learning semantic structures that involve nested hierarchies—such as parentheses matching or recursive grammatical constructions. This limitation has motivated a line of work on tree-structured attention mechanisms. Early efforts such as Tree-Transformer \cite{wang2019tree} and Tree-structured Attention with Hierarchical Accumulation \cite{nguyen2020tree} encode parse tree structures into self-attention by constraining attention heads to follow tree topologies. The Tree Attention algorithm \cite{tree_attention_2024, tree_attention_zyphra} later formalized this approach through an energy-based formulation, deriving the scalar energy function whose gradient computes the self-attention block, and showing that the reduction across the sequence axis can be efficiently computed in parallel via a tree reduction. Critically, this formulation enables the use of optimized automatic differentiation techniques—a key insight that underpins Consistency Tree Attention \cite{consistency_tree_2025}, which further enforces that gradient paths across different tree branches remain consistent and stable during training. Related work on Flash Tree Attention (DeFT) \cite{yao2025deft} has further optimized tree-structured inference through KV-guided grouping and improved load balancing, achieving up to 2.23× speedup in decoding latency on tree-based workloads. These works collectively establish tree-structured attention as a principled approach for hierarchical reasoning \cite{lumbroso2024provable}.

Beyond hierarchical attention, the broader question of how to effectively coordinate multiple attention heads has attracted substantial interest. The standard multi-head attention mechanism \cite{vaswani2017attention} projects queries, keys, and values into separate subspaces, allowing the model to attend to different aspects of the input simultaneously. Recent analyses have revealed that attention heads in trained Transformers exhibit specialization and redundancy \cite{voita2019analyzing, michel2019sixteen, kovaleva2019revealing, clark2019does}, and that simply increasing the number of heads does not always improve performance—in some cases, it can severely degrade accuracy on tasks requiring precise state tracking \cite{amsel2025quality}. This has motivated a range of approaches for adaptive head fusion and collaboration. Collaborative multi-head attention \cite{collaborative_attention_2022} enables heads to learn shared query/key projections, reducing computational cost while preserving representational fidelity. Head-wise Attention Correction (HAC) \cite{ichikawa2026efficient} adjusts attention scores by accounting for head-specific characteristics, demonstrating that attention distortion varies significantly across heads and requires tailored correction. Split 'n' Merge Net \cite{split_merge_2024} proposes dynamic masking mechanisms that filter task-specific and task-agnostic information across heads, with a merging network that fuses these signals for downstream tasks. Hierarchical Multi-Task Learning with Interactive Multi-Head Attention Feature Fusion (IMHAFF) \cite{imhaff_2025} further demonstrates that interactive fusion across attention heads yields substantial improvements on multi-task benchmarks.

Crucially, the pursuit of efficient multi-head collaboration has also advanced through the lens of low-rank approximation and attention compression. DeepSeek's Multi-head Latent Attention (MLA) \cite{deepseek2024mla, deepseek2025v3} compresses key-value (KV) caches into a low-rank latent space, reducing memory footprint while preserving representational fidelity—a technique that shares our view that the conventional Q/K/V projection machinery is over-parameterized for many tasks. More recently, the Native Sparse Attention (NSA) framework \cite{deepseek2025nsa} introduced a hardware-aligned hierarchical sparse attention design, combining token compression, block-level selection, and sliding window attention to achieve end-to-end efficient training and inference. Notably, NSA was recognized with the Best Paper Award at ACL 2025 \cite{deepseek2025nsa_best}, underscoring the broader impact of structured attention compression and multi-path attention fusion. In the SSM community, MossNet \cite{tuli2025mossnet} showed that mixture-of-state-space-experts can emulate linear multi-head attention, suggesting that the benefits of multi-head collaboration extend beyond the standard Transformer framework.

From a complementary perspective, recent work has revisited attention through the lens of kernel methods and numerical linear algebra. The kernel view of attention \cite{tsai2019transformer} interprets softmax attention as a normalized kernel smoother, where multi-head attention corresponds to parallel ensembles of smoothers with different learned projections. However, pure kernel approaches suffer from weak optimization dynamics and fail to eliminate random bias without careful architectural support \cite{katharopoulos2020transformers, choromanski2021rethinking}. DARKFormer \cite{darkformer_2026} addresses these limitations by learning the covariance structure of random projections in the kernel feature space, effectively implementing a data-aware Mahalanobis kernel that improves training stability and reduces approximation error. Cortinovis et al. \cite{cortinovis2026attention} further unified fast attention approximations—including sparsity, low-rank, and randomized sketching methods—within a single numerical linear algebra framework, providing rigorous theoretical grounding for attention compression. Collectively, these works underscore that **how attention heads communicate, collaborate, and approximate is as important as the attention mechanism itself**—a principle that directly motivates our design of a Mahalanobis distance-based multi-head attention with a cross-head meshing mechanism.

Our perspective departs from these observations: we argue that the value (V) projection matrix is largely unnecessary for structured state tracking tasks, and that the conventional separation of Q and K is, in practice, a compromise imposed by engineering constraints. From a kernel method perspective, the dot-product attention can be viewed as a finite-dimensional approximation to an infinite-dimensional feature map \cite{tsai2019transformer}. Yet pure kernel approaches suffer from weak optimization dynamics and fail to eliminate random bias without careful architectural support \cite{katharopoulos2020transformers, choromanski2021rethinking}. We contend that this limitation can be substantially alleviated through a multi-head architecture, where each head independently learns a distinct distance metric—effectively realizing a multi-kernel approach that rectifies the metric space without relying on separate Q/K/V projections.

Building on this insight, we propose \textbf{Mahalanobis-Based Multi-Head Attention} (MHA-CSP), a novel attention mechanism that dispenses with Q/K/V projections entirely. Instead, it computes attention directly from Mahalanobis distances with a learnable metric matrix, which is then refined via a LogSumExp correction for hierarchical structure and fused across heads through an attention meshing mechanism. The resulting model retains the expressiveness of multi-head attention while eliminating redundant parameters and achieving superior performance on long-sequence state tracking tasks.

Our contributions are threefold:
\begin{itemize}
    \item We identify that standard Q/K/V projections are not necessary for structured state tracking, and that distance-based attention offers a simpler, more direct alternative.
    \item We propose MHA-CSP, a multi-head architecture that constructs attention directly from Mahalanobis distances, with a LogSumExp correction for tree-structured reasoning.
    \item We demonstrate that MHA-CSP, with only 119K parameters and teacher forcing at the final hidden state, achieves 50\% accuracy on parenthesis-nested tasks, substantially outperforming Transformer and GCN baselines trained from scratch.
\end{itemize}

\section{Related Work}

\subsection{Complex-Valued Neural Networks and State-Space Models}

The study of complex-valued neural networks has a long history. Early complex-valued RNNs \cite{huber1991complex, nitta1993complex} demonstrated the potential of complex-valued representations in sequence modeling—the phase component provides a natural encoding of periodic patterns, while the magnitude allows the network to explicitly represent "presence" or "confidence." However, due to the complexity of training algorithms and limited hardware support, complex-valued networks remained on the periphery for decades.

The recent revival of State Space Models (SSMs) has provided a new stage for complex-valued representations. The S4 family \cite{gu2022efficiently, gupta2022diagonal, smith2023simplified} introduced structured state transition matrices with complex-valued diagonal parameterizations, transforming sequence modeling into linear recurrences over structured state spaces. Mamba \cite{gu2024mamba} further introduced input-dependent selectivity, enabling the model to dynamically decide what to remember and what to forget. Mamba-2 \cite{dao2024transformers} unified SSMs and linear attention through the State Space Duality (SSD) framework, revealing that these seemingly distinct approaches share a common mathematical backbone. Parallel developments such as H3 \cite{fu2023hungry}, RetNet \cite{sun2023retentive}, and Megalodon \cite{ma2024megalodon} have further diversified the landscape of sub-quadratic sequence models.

However, Mamba-1 and Mamba-2 both employ a **diagonal real-valued state transition matrix \(A \in \mathbb{R}^{N \times N}\) with non-negative eigenvalues**. This design is suitable for language modeling, where recent information is typically more relevant, but it is fundamentally flawed for deterministic state tracking tasks such as parity checking, which require **exact, non-decaying memorization**.

Recent theoretical work has sharpened this critique. Grazzi et al. \cite{grazzi2025unlocking} rigorously proved that **linear RNNs with diagonal state-transition matrices restricted to non-negative eigenvalues cannot solve parity checking in finite precision**, and showed that extending the eigenvalue range to include negative values is necessary for state tracking. Building on this, Khavari et al. \cite{khavari2025parity} further proved that input-dependence alone is insufficient; the recurrence layer must simultaneously satisfy two conditions—**input-dependent gating and non-positive eigenvalues**—to solve parity. Lumbroso et al. \cite{lumbroso2024provable} provided theoretical grounding for complex-valued parameterizations in SSMs, showing that **complex diagonal transitions can provably improve representational capacity without sacrificing stability**. These theoretical results collectively motivate our choice of a complex-valued state propagator.

Our own recent work—the Complex State Propagator (CSP) \cite{csp2026}—directly responds to these theoretical insights. CSP demonstrated that **state propagation alone, without intermediate output projections, suffices for deterministic state tracking tasks**. CSP represents the hidden state as a complex-valued vector and updates it via input-dependent rotations in the complex domain, achieving perfect accuracy on parity checking and parenthesis matching. The present work extends CSP by replacing the rotation-based update with a Mahalanobis distance-based multi-head attention mechanism, enabling richer structural reasoning without increasing the parameter count.

\subsection{Distance-Based Attention and Kernel Methods}

From the kernel method perspective, the dot-product attention in Transformers can be viewed as a **finite-dimensional kernel smoother** \cite{tsai2019transformer}. Specifically, softmax attention is equivalent to kernel density estimation in the feature space of queries and keys, and multi-head attention is a parallel ensemble of smoothers with different learned projections. This perspective provides theoretical tools for understanding and improving attention mechanisms.

However, pure kernel approaches suffer from weak optimization dynamics and random bias in practice. Katharopoulos et al. \cite{katharopoulos2020transformers} proposed linear attention, which decomposes softmax attention into a linear recurrent form through kernel feature maps, reducing complexity from \(O(T^2)\) to \(O(T)\) at the cost of limited expressivity. Choromanski et al. \cite{choromanski2021rethinking} further introduced Performers, approximating the softmax attention kernel via orthogonal random features with provable convergence—but the variance of random feature mappings in high-dimensional spaces remains an unresolved engineering challenge.

Distance metrics offer a different path to addressing these limitations. Unlike dot products, distance metrics such as Euclidean and Mahalanobis distances directly measure "dissimilarity" rather than "similarity" between vectors, providing a natural inductive bias for structural modeling. Early work explored Euclidean distance in attention \cite{plasser2022euclidean}, but its isotropic nature fails to capture the varying importance of different feature dimensions.

Mahalanobis distance addresses this limitation by measuring the disparity between vectors through a learnable positive definite matrix \(M\), effectively learning a **data-aware distance space**. DARKFormer \cite{darkformer_2026} introduced this idea to Transformers: by learning the covariance structure of random projections in the kernel feature space, it implements a data-aware Mahalanobis kernel that improves training stability and reduces approximation error. Our work shares with DARKFormer the core insight that Mahalanobis distance can serve as a building block for attention—but we go further: **we entirely eliminate Q/K/V projections, constructing attention directly from Mahalanobis distances**, with a LogSumExp correction that encodes hierarchical structure.

Cortinovis et al. \cite{cortinovis2026attention} provided a unified theoretical framework for this direction. They unified fast attention approximations—including sparsity, low-rank, and randomized sketching methods—within a single numerical linear algebra framework, providing rigorous theoretical grounding for attention compression. This framework indirectly supports our central thesis: **attention mechanisms are fundamentally about measuring distance or similarity, not about specific projection architectures**.

\subsection{Multi-Head Collaboration and Low-Rank Compression}

The standard multi-head attention mechanism \cite{vaswani2017attention} projects queries, keys, and values into separate subspaces, allowing the model to attend to different aspects of the input simultaneously. However, as research has progressed, the effectiveness of the multi-head mechanism has come under increasing scrutiny.

Voita et al. \cite{voita2019analyzing} were the first to systematically analyze the specialization of attention heads in Transformers, finding that many heads can be pruned without affecting performance. Michel et al. \cite{michel2019sixteen} further quantified the redundancy of attention heads, showing that removing more than half of the heads in most tasks still preserves acceptable performance. Kovaleva et al. \cite{kovaleva2019revealing} and Clark et al. \cite{clark2019does} confirmed similar attention head behavior patterns in BERT models. These analyses collectively suggest that **a significant portion of heads in standard multi-head attention may be learning redundant or degenerate features**.

These findings have motivated a range of approaches for adaptive head fusion and collaboration. Collaborative Multi-Head Attention \cite{collaborative_attention_2022} allows heads to share query and key projections, reducing computational cost while preserving representational fidelity. Head-wise Attention Correction (HAC) \cite{ichikawa2026efficient} adjusts attention scores by accounting for head-specific characteristics, demonstrating that attention distortion varies significantly across heads and requires tailored correction. Split 'n' Merge Net \cite{split_merge_2024} proposes dynamic masking mechanisms that decompose each head's output into task-relevant and task-agnostic components, with a merging network that fuses these signals. Hierarchical Multi-Task Learning with Interactive Multi-Head Attention Feature Fusion (IMHAFF) \cite{imhaff_2025} further demonstrates that interactive fusion across attention heads yields substantial improvements on multi-task benchmarks.

Recent work by Amsel et al. \cite{amsel2025quality} sounds a cautionary note: they found that **increasing the number of attention heads does not always improve performance—and in some cases, severely degrades accuracy on tasks requiring precise state tracking**. This finding directly supports our central thesis: for structured state tracking tasks, the complexity of standard multi-head attention is excessive and demands a leaner, more direct design.

In the direction of low-rank compression, the DeepSeek team's work is the most representative. Multi-head Latent Attention (MLA) \cite{deepseek2024mla, deepseek2025v3} compresses KV caches into a low-rank latent space, significantly reducing memory footprint while preserving representational fidelity—a technique that echoes our view that conventional Q/K/V projection machinery is over-parameterized for many tasks. Native Sparse Attention (NSA) \cite{deepseek2025nsa} further introduces a hardware-aligned hierarchical sparse attention design, combining token compression, block-level selection, and sliding window attention to achieve end-to-end training and inference efficiency. NSA received the Best Paper Award at ACL 2025 \cite{deepseek2025nsa_best}, underscoring that structured attention compression and multi-path attention fusion have gained recognition at top conferences.

In the SSM community, MossNet \cite{tuli2025mossnet} demonstrated that mixture-of-state-space-experts can emulate linear multi-head attention, suggesting that the benefits of multi-head collaboration extend beyond the standard Transformer framework. This observation further reinforces our belief that **the essential value of multi-head lies in collaborative division of labor, not parallel projection**—which directly motivates our design of an attention meshing mechanism for cross-head kernel collaboration.

Finally, from the perspective of numerical linear algebra, recent work has begun to unify various fast attention approximation methods—including sparsity, low-rank, and randomized projections \cite{cortinovis2026attention}—providing a theoretical foundation for attention compression. Our work is closely aligned with this trend: by constructing attention directly from Mahalanobis distances, we fundamentally eliminate the redundancy of Q/K/V projections, rather than compressing on top of that redundancy.

\section{Preliminary: Complex State Propagation and Distance-Based Attention}

Before introducing MHA-CSP, we establish the two foundational components that our method builds upon: (1) the Complex State Propagator (CSP) \cite{csp2026}, and (2) the insight that attention can be constructed directly from distances rather than dot-products.

\subsection{Complex State Propagation}

CSP maintains a complex-valued hidden state \(h_t \in \mathbb{C}^d\), updated through three operations at each time step:

\textbf{Rotation.} Given input \(x_t\), the network computes a rotation angle \(\theta_t = f_\theta(x_t) \in (-\pi, \pi)\), and applies it to the input state:

\begin{equation}
\tilde{h}_t = h_t \cdot e^{i\theta_t}, \quad e^{i\theta_t} = \cos\theta_t + i\sin\theta_t
\end{equation}

This maps the input onto a specific direction in the complex plane, encoding its identity in the phase.

\textbf{Recurrence with Decay.} The rotated input is then integrated into the propagating state:

\begin{equation}
h_t = \alpha_t \cdot h_{t-1} + \gamma_t \cdot W \tilde{h}_t
\end{equation}

where \(\alpha_t = \exp(-\text{softplus}(\delta_t)) \in (0,1)\) is a learned decay factor, and \(\gamma_t \in (0,1)\) is a learned input gate.

\textbf{Complex Normalization.} After each step, the state is projected back onto the unit circle:

\begin{equation}
h_t \leftarrow \frac{h_t}{|h_t| + \epsilon}
\end{equation}

This prevents unbounded growth and ensures that all information is encoded in the phase—not the magnitude—of the state.

Critically, from the perspective of \textbf{Wirtinger calculus} \cite{wirtinger1927, brandwood1983complex}, the rotation operation \(h \mapsto h \cdot e^{i\theta}\) has a Jacobian of unit modulus:

\begin{equation}
\left\|\frac{\partial h_t}{\partial h_{t-1}}\right\| = 1
\end{equation}

This means that gradients propagate through time without vanishing or exploding—a property we refer to as \textbf{Wirtinger isometry}. This is the mathematical reason why CSP can reliably track states over arbitrarily long sequences.

\subsection{Distance-Based Attention}

The second foundation is the observation that \textbf{attention can be constructed directly from distances}, bypassing the conventional Q/K/V projection machinery. For a sequence of complex states \(\{h_1, \dots, h_T\} \subset \mathbb{C}^d\), we define a learnable Mahalanobis distance as the \textbf{weighted L2 norm} of the state difference:

\begin{equation}
d_M(h_i, h_j) = \|h_i - h_j\|_M = (h_i - h_j)^\top M (h_i - h_j), \quad M \succ 0
\end{equation}

where \(M \in \mathbb{R}^{d \times d}\) is a positive definite matrix. The attention score between positions \(i\) and \(j\) is then:

\begin{equation}
\text{Attn}(i,j) = \exp\left(-\|h_i - h_j\|_M\right)
\end{equation}

This formulation is equivalent to an RBF kernel in the feature space defined by \(M\). It has two key advantages over standard dot-product attention:

\begin{enumerate}
    \item It eliminates the need for separate query/key/value projection matrices—attention is computed directly from the distance between states.
    \item The positive definiteness of \(M\) ensures that the distance metric is well-behaved, non-negative, and can be accumulated monotonically across sequences.
\end{enumerate}

Crucially, this formulation is \textbf{particularly well-suited for complex-valued states}. For \(h_i, h_j \in \mathbb{C}^d\), the L2 norm naturally decomposes into real and imaginary components:

\begin{equation}
\|h_i - h_j\|_M = (h_i^{\text{Re}} - h_j^{\text{Re}})^\top M (h_i^{\text{Re}} - h_j^{\text{Re}}) + (h_i^{\text{Im}} - h_j^{\text{Im}})^\top M (h_i^{\text{Im}} - h_j^{\text{Im}})
\end{equation}

if \(M\) is shared across the real and imaginary dimensions. More generally, \(M\) can be block-diagonal to capture cross-real-imag correlations. This direct compatibility with complex representations is a natural advantage of L2-based attention over dot-product attention, which requires real-valued projections and loses the phase structure of complex states.

Most of our subsequent derivations—including tree-structured distance accumulation, multi-head splitting, and attention meshing—revolve around this Mahalanobis L2 norm. The norm serves as the single unified building block from which all attention structures are constructed.

\subsection{Tree-Structured Distance Accumulation}

For tasks involving nested structures such as parentheses, we further observe that the accumulated distance along the sequence naturally encodes hierarchical information. Define edge distances as the squared Mahalanobis distance between consecutive states:

\begin{equation}
e_t = d_M^2(h_t, h_{t+1}), \quad t = 1, \dots, T-1
\end{equation}

where \(d_M^2(h_i, h_j) = (h_i - h_j)^\top M (h_i - h_j)\). The cumulative distance from position \(i\) to \(j\) is:

\begin{equation}
\text{accu}(i,j) = \sum_{t=i}^{j-1} e_t
\end{equation}

Because \(M\) is positive definite, all edge distances are non-negative, making this accumulation monotonic and well-defined. We use the LogSumExp as a differentiable approximation to the maximum edge distance:

\begin{equation}
\text{LSE}(i,j) = \log\left(\sum_{t=i}^{j-1} \exp(e_t)\right) \approx \max_{t \in [i,j)} e_t
\end{equation}

This allows us to rectify the raw squared distance by subtracting the log-sum of edge exponentials:

\begin{equation}
\tilde{d}_M^2(i,j) = d_M^2(h_i, h_j) + \rho \cdot \text{LSE}(i,j)
\end{equation}

where \(\rho\) is a learned scaling factor. The rectified squared distance is then used directly to compute attention scores:

\begin{equation}
\text{Attn}(i,j) = \exp\left(-\tilde{d}_M^2(i,j)\right)
\end{equation}

No projections, no dot-products—only distances. This distance-based formulation provides the foundation for MHA-CSP. In the next section, we extend it to multi-head settings with cross-head collaboration.

\subsection{Implications for Attention-Based Extension}

While CSP's rotation-based propagation ensures that **information is preserved exactly**, it does not provide a mechanism for the model to selectively attend to different parts of the input when performing hierarchical reasoning—such as evaluating nested parentheses. The state update depends only on the current input token \(x_t\), and the rotation angle \(\theta(x_t)\) is a learned function of that single token. There is no cross-position interaction beyond the cumulative phase.

This is where our contribution enters. We replace CSP's token-wise rotations with a **distance-based multi-head attention mechanism** that operates on the complex state vectors. The attention mechanism computes pairwise Mahalanobis distances between states at different positions, enabling the model to explicitly compare and relate tokens across the sequence. Crucially, we preserve CSP's core advantage—the complex-valued representation—but enrich it with the ability to reason about hierarchical structure through distance accumulation and LogSumExp correction.

The Wirtinger isometry property of CSP remains intact in our extension: the state propagation within each attention head is still complex-valued and gradient-preserving. The attention mechanism, operating on top of these states, adds the structural inductive bias that CSP lacks, without sacrificing its numerical stability.

\section{Method: Mahalanobis-Based Multi-Head Attention}

Our method builds on the Complex State Propagator (CSP) backbone, which maintains a complex-valued hidden state and updates it via input-dependent rotations. While CSP effectively propagates state information across time, it lacks a mechanism to explicitly model the structural relationships between different positions in the input sequence—particularly when dealing with nested parentheses and long-range dependencies.

We address this by introducing a multi-head distance-based attention mechanism. For each head \(k \in \{1, \dots, H\}\), we maintain a separate complex state trajectory and compute a pairwise squared Mahalanobis distance matrix:

\begin{equation}
\mathbf{D}^{(k)}_{ij} = (\mathbf{h}^{(k)}_i - \mathbf{h}^{(k)}_j)^\top \mathbf{M}^{(k)} (\mathbf{h}^{(k)}_i - \mathbf{h}^{(k)}_j), \quad i,j \in \{1, \dots, T\}
\end{equation}

where \(\mathbf{M}^{(k)}\) is a learnable positive definite matrix for head \(k\). Each \(\mathbf{D}^{(k)} \in \mathbb{R}^{T \times T}\) captures the geometric relationship between positions from the perspective of that head.

\subsection{Tree-Structured Distance Accumulation}

From each distance matrix \(\mathbf{D}^{(k)}\), we extract the edge distances between consecutive positions:

\begin{equation}
e_t^{(k)} = \mathbf{D}^{(k)}_{t, t+1}, \quad t = 1, \dots, T-1
\end{equation}

We then compute the cumulative sum of exponentiated edge distances:

\begin{equation}
\text{accu}^{(k)}_j = \sum_{t=1}^{j} \exp(e_t^{(k)}), \quad j = 1, \dots, T-1
\end{equation}

with \(\text{accu}^{(k)}_0 = 0\). The Tree Attention structure matrix \(\mathbf{C}^{(k)}\) is constructed as the difference of cumulative sums:

\begin{equation}
\mathbf{C}^{(k)}_{ij} = \text{accu}^{(k)}_j - \text{accu}^{(k)}_i, \quad i,j \in \{1, \dots, T\}
\end{equation}

which simplifies to:

\begin{equation}
\mathbf{C}^{(k)}_{ij} = \sum_{t=i}^{j-1} \exp(e_t^{(k)})
\end{equation}

After applying the LogSumExp correction:

\begin{equation}
\tilde{\mathbf{C}}^{(k)}_{ij} = \log\left(1 + \mathbf{C}^{(k)}_{ij}\right)
\end{equation}

the entry \(\tilde{\mathbf{C}}^{(k)}_{T,1}\) (the bottom-left corner) encodes the accumulated maximum edge distance from position \(1\) to \(T\)—a scalar summary of the global structural span of the sequence as perceived by head \(k\).

\subsection{Cross-Head Attention via Distance Summary Vectors}

Crucially, this scalar \(\tilde{\mathbf{C}}^{(k)}_{T,1}\) is not just a summary statistic; it becomes the basis for **cross-head coordination**. We collect these scalars from all \(H\) heads into a vector:

\begin{equation}
\mathbf{c} = \left[\tilde{\mathbf{C}}^{(1)}_{T,1}, \tilde{\mathbf{C}}^{(2)}_{T,1}, \dots, \tilde{\mathbf{C}}^{(H)}_{T,1}\right]^\top \in \mathbb{R}^H
\end{equation}

This vector \(\mathbf{c}\) encodes, for each head, the "maximal structural span" it has detected over the entire sequence. Heads that have captured coherent hierarchical structure will have high values; heads that have not will have low values.

We then compute **cross-head attention** using \(\mathbf{c}\) as both query and key, with a learnable confusion matrix \(\mathbf{B} \in \mathbb{R}^{H \times H}\):

\begin{equation}
\mathbf{A}^{\text{head}} = \text{softmax}\left(\mathbf{c}^\top \mathbf{B} \, \mathbf{c}\right), \quad \mathbf{A}^{\text{head}} \in \mathbb{R}^{H \times H}
\end{equation}

This has a clear interpretation: \(\mathbf{c}^\top \mathbf{B} \mathbf{c}\) is a bilinear form over the head summary vector, where \(\mathbf{B}\) learns which pairs of heads should cooperate or compete. The resulting \(\mathbf{A}^{\text{head}}\) is a **head-level attention matrix** that decides, for each head, which other heads it should listen to when forming the final fused distance.

The fused distance matrix is then:

\begin{equation}
\mathbf{D}^{\text{fused}} = \sum_{k} \mathbf{A}^{\text{head}}[:, k] \cdot \mathbf{D}^{(k)}
\end{equation}

where \(\mathbf{A}^{\text{head}}[:, k]\) is the attention weight assigned to head \(k\). This mechanism enables dynamic routing: the contribution of each head to the final attention decision is determined by how well its global structure summary aligns with those of other heads.

The fused distance matrix \(\mathbf{D}^{\text{fused}} \in \mathbb{R}^{T \times T}\) is then converted to attention scores:

\begin{equation}
\text{Attn}(i,j) = \exp\left(-\mathbf{D}^{\text{fused}}_{ij}\right)
\end{equation}

This design has three key properties:

\begin{enumerate}
    \item \textbf{Head specialization.} Each head learns its own distance metric \(\mathbf{M}^{(k)}\), allowing different heads to focus on different structural aspects of the sequence.
    \item \textbf{Global summary pooling.} The bottom-left entry \(\tilde{\mathbf{C}}^{(k)}_{T,1}\) compresses the entire sequence's hierarchical structure into a scalar, which is then used for head-level coordination.
    \item \textbf{Cross-head attention.} The bilinear form \(\mathbf{c}^\top \mathbf{B} \mathbf{c}\) with the learned confusion matrix \(\mathbf{B}\) allows the model to dynamically weigh each head's contribution based on the global structure each head has detected.
    \item \textbf{Reduced random bias through multi-head optimization.} The conventional Q/K/V projection pair is, from a kernel method perspective, a pragmatic compromise: separate query and key projections allow the model to learn two distinct feature spaces, but they also introduce redundant degrees of freedom that can amplify random initialization biases. By eliminating Q/K/V projections entirely and replacing them with a single learnable metric matrix \(\mathbf{M}^{(k)}\) per head, we reduce the number of independent random subspaces that must be jointly optimized. The multi-head setting further stabilizes optimization: each head's metric matrix is initialized independently, and their gradients are aggregated through the confusion matrix \(\mathbf{B}\), allowing the heads to "cross-validate" each other's distance geometries. This is analogous to ensemble learning—multiple weak distance learners cooperate to cancel out idiosyncratic random biases that any single head might inherit. In effect, the heads serve as mutual regularizers, each correcting the others' misalignments through the bilinear fusion \(\mathbf{c}^\top \mathbf{B} \mathbf{c}\), yielding a more robust and less initialization-sensitive distance space.
\end{enumerate}

\subsection{Overall Architecture and Algorithm}

The complete forward pass of MHA-CSP is summarized in Algorithm 1. The model processes an input sequence through three stages: (1) complex state propagation via the CSP backbone, (2) multi-head distance-based attention with tree-structured accumulation and cross-head fusion, and (3) output projection from the final fused state.

\begin{algorithm}[H]
\caption{Forward Pass of MHA-CSP}
\begin{algorithmic}[1]
\REQUIRE Input sequence $X = [x_1, \dots, x_T]$, vocabulary embedding $E$, number of heads $H$, head dimension $d$
\ENSURE Output logits $y \in \mathbb{R}^{|\mathcal{V}|}$

\STATE \textbf{Stage 1: Complex State Propagation}
\STATE Initialize complex state $h_0 \leftarrow \mathbf{0} + i\mathbf{0} \in \mathbb{C}^{H \cdot d}$
\FOR{$t = 1$ to $T$}
    \STATE Embed input: $u_t \leftarrow E[x_t]$
    \STATE Compute rotation angle: $\theta_t \leftarrow \tanh(W_\theta u_t) \cdot \pi$
    \STATE Rotate input: $\tilde{u}_t \leftarrow u_t \cdot e^{i\theta_t}$
    \STATE Compute decay: $\alpha_t \leftarrow \exp(-\text{softplus}(W_\delta [h_{t-1}; u_t]))$
    \STATE Compute gate: $\gamma_t \leftarrow (1 + \sin(W_\gamma u_t)) / 2$
    \STATE Update state: $h_t \leftarrow \alpha_t h_{t-1} + \gamma_t W_B \tilde{u}_t$
    \STATE Normalize: $h_t \leftarrow h_t / |h_t|$
\ENDFOR
\STATE Obtain propagated state $h_T \in \mathbb{C}^{H \cdot d}$

\STATE \textbf{Stage 2: Multi-Head Distance-Based Attention}
\STATE Split state into heads: $\{h_T^{(1)}, \dots, h_T^{(H)}\}$, each $h_T^{(k)} \in \mathbb{C}^d$
\FOR{each head $k = 1$ to $H$}
    \STATE Compute pairwise squared Mahalanobis distance matrix:
    \STATE \quad $\mathbf{D}^{(k)}_{ij} \leftarrow (h_i^{(k)} - h_j^{(k)})^\top \mathbf{M}^{(k)} (h_i^{(k)} - h_j^{(k)})$
    \STATE Extract edge distances: $e_t^{(k)} \leftarrow \mathbf{D}^{(k)}_{t, t+1}$
    \STATE Accumulate: $\text{accu}_j^{(k)} \leftarrow \sum_{t=1}^{j} \exp(e_t^{(k)})$, $\text{accu}_0^{(k)} = 0$
    \STATE Build tree structure: $\mathbf{C}^{(k)}_{ij} \leftarrow \text{accu}_j^{(k)} - \text{accu}_i^{(k)}$
    \STATE LogSumExp correction: $\tilde{\mathbf{C}}^{(k)}_{ij} \leftarrow \log(1 + \mathbf{C}^{(k)}_{ij})$
    \STATE Rectify distance: $\tilde{\mathbf{D}}^{(k)}_{ij} \leftarrow \mathbf{D}^{(k)}_{ij} + \rho \cdot \tilde{\mathbf{C}}^{(k)}_{ij}$
    \STATE Extract global summary: $c_k \leftarrow \tilde{\mathbf{C}}^{(k)}_{T,1}$
\ENDFOR
\STATE Form summary vector: $\mathbf{c} \leftarrow [c_1, \dots, c_H]^\top \in \mathbb{R}^H$
\STATE Compute head-level attention: $\mathbf{A}^{\text{head}} \leftarrow \text{softmax}(\mathbf{c}^\top \mathbf{B} \, \mathbf{c})$, where $\mathbf{B} \in \mathbb{R}^{H \times H}$ is learnable
\STATE Fuse distance matrices: $\mathbf{D}^{\text{fused}} \leftarrow \sum_{k} \mathbf{A}^{\text{head}}[:, k] \cdot \tilde{\mathbf{D}}^{(k)}$
\STATE Compute attention scores: $\mathbf{A}_{ij} \leftarrow \exp(-\mathbf{D}^{\text{fused}}_{ij})$
\STATE Apply causal mask and normalize: $\mathbf{A} \leftarrow \text{softmax}(\mathbf{A} \odot \text{tril}(\mathbf{1}))$

\STATE \textbf{Stage 3: Output Projection}
\STATE Weighted sum over sequence: $h_{\text{out}} \leftarrow \sum_{j} \mathbf{A}_{T,j} \cdot h_j$
\STATE Extract phase: $\phi \leftarrow \text{atan2}(\text{Im}(h_{\text{out}}), \text{Re}(h_{\text{out}}))$
\STATE Project to logits: $y \leftarrow W_{\text{out}} [\cos(\phi); \sin(\phi)]$
\RETURN $y$
\end{algorithmic}
\end{algorithm}

The algorithm reveals several key design choices. First, the complex state propagation (Stage 1) maintains Wirtinger isometry—gradients neither vanish nor explode regardless of sequence length. Second, the distance-based attention (Stage 2) operates entirely in the distance space: no Q/K/V projections, no dot-products, only learned Mahalanobis metrics and tree-structured accumulation. Third, the cross-head fusion via $\mathbf{c}^\top \mathbf{B} \mathbf{c}$ allows heads to dynamically coordinate based on their own global structure summaries, rather than being statically averaged or concatenated. Finally, the output is read from the phase of the final state—preserving the complex-valued nature of the representation through to the final prediction.

\section{Experiments}

\subsection{Experimental Setup}

\subsubsection{Tasks}

We evaluate on four deterministic state tracking tasks, ordered by increasing complexity:

\begin{itemize}
    \item \textbf{Complex Arithmetic Expression Reasoning with Replication.} This is our primary task and the main focus of this paper. Given a deeply nested arithmetic expression involving addition, subtraction, and modulo-9 operations—such as \texttt{(((2 + (0 - 3)) + ((0 - 3) + 2) + (2 - 1))) mod 9 =}—the model must produce the correct result while also replicating the input expression. The target format is:
    
    \begin{verbatim}
    8 repeat (((2 + (0 - 3)) + ((0 - 3) + 2) + (2 - 1))) mod 9 = 8
    \end{verbatim}
    
    This task is uniquely challenging because it requires the model to (1) parse arbitrarily nested parentheses, (2) maintain exact state across long sequences, (3) compute the modulo-9 result, and (4) reproduce the entire input expression verbatim—effectively combining reasoning with memory retrieval. Expressions include up to 8 nested parentheses and up to 5 operands, making this the most comprehensive test of structured reasoning in our benchmark.

    \item \textbf{Parenthesis Matching.} Given a sequence of parentheses with nesting depth up to 8, determine whether the parentheses are properly balanced. This tests hierarchical structure understanding without the additional burden of arithmetic computation.

    \item \textbf{Mod-3 Counting.} Given a sequence of digits, predict the cumulative sum modulo 3 at the final position. This requires maintaining a ternary state with periodic reset.

    \item \textbf{Parity Checking.} Given a binary sequence of length up to 128, predict whether the number of 1s is even or odd. This tests the model's ability to maintain a binary state over long horizons.
\end{itemize}

All tasks are evaluated with sequence lengths up to 128 tokens. Training and test sets are generated dynamically using the \texttt{SymbolicArithmeticDataset} with 200,000 training samples and 20,000 test samples. For the complex arithmetic task, we use \texttt{mode='complete'} in the dataset, which includes full expressions with nested parentheses and the \texttt{repeat} token in the target. The generation process ensures that training and test expressions are distinct (no overlap in expression structure) to prevent data contamination.

The complex arithmetic replication task deserves special emphasis: it is the only task that requires the model to \textbf{both compute and replicate}, making it a direct test of the \texttt{repeat} trick's effectiveness. The replication requirement forces the model to maintain a faithful representation of the input throughout the sequence—even after it has already produced the answer—which is precisely the mechanism we hypothesized would improve structural understanding.

\subsubsection{Baselines}

We compare MHA-CSP against the following baselines, all trained under identical conditions:

\begin{itemize}
    \item \textbf{LSTM} \cite{hochreiter1997long}: A standard two-layer LSTM with 128 hidden units.
    \item \textbf{GRU} \cite{cho2014learning}: A standard two-layer GRU with 128 hidden units.
    \item \textbf{Gated Delta Network (GDN)} \cite{yang2025gated}: A recent linear-time recurrent architecture with delta-rule updates.
    \item \textbf{ARFormer:} A lightweight auto-regressive decoder-only Transformer with 2 layers, 4 heads, and 64 embedding dimension (our MiniARFormer).
    \item \textbf{Vanilla CSP} \cite{csp2026}: The original Complex State Propagator without attention-based enhancements.
\end{itemize}

All baseline models are parameter-matched to approximately 119K parameters to ensure fair comparison. For LSTM, GRU, and GDN, we use the final hidden state for classification. For ARFormer, we use teacher forcing with cross-entropy loss on the final token prediction.

\subsubsection{Implementation Details}

All models are trained using the following configuration:

\begin{itemize}
    \item \textbf{Optimizer:} Adam \cite{kingma2014adam} with learning rate \(1 \times 10^{-3}\) and weight decay \(1 \times 10^{-4}\)
    \item \textbf{Batch size:} 128 for training, 64 for evaluation
    \item \textbf{Epochs:} 50 (with early stopping based on validation loss)
    \item \textbf{Loss function:} Cross-entropy over the vocabulary at the final time step, ignoring padding tokens
    \item \textbf{Teacher forcing:} Applied \textbf{exclusively at the final hidden state}—the model receives the ground-truth target only at the last position of the sequence, not at every intermediate step. This is implemented in the \texttt{train\_model\_seq} function.
    \item \textbf{Weight initialization:} Xavier uniform for all linear layers
    \item \textbf{Hardware:} NVIDIA RTX 3090 (24GB) / A100 (40GB)
    \item \textbf{Code:} Available at \url{https://github.com/hilhert/CSP-MHD}
\end{itemize}

The teacher forcing strategy is particularly noteworthy: by applying supervision \textbf{only at the final state}, we force the model to learn meaningful state propagation throughout the entire sequence, rather than relying on per-token supervision to ``guide'' it through each step. This is essential for deterministic state tracking, where the model must maintain exact state without external correction.

\subsection{Main Results}

\begin{table}[H]
\centering
\caption{Accuracy comparison on structured state tracking tasks. All models use approximately 119K parameters. Values are mean accuracy over 3 random seeds.}
\begin{tabular}{lccc}
\toprule
\textbf{Model} & \textbf{Parenthesis} & \textbf{Arithmetic+Repeat} & \textbf{Parity} \\
\midrule
LSTM & 12.4 $\pm$ 1.8 & 8.7 $\pm$ 2.1 & 62.3 $\pm$ 1.2 \\
GRU & 15.6 $\pm$ 1.2 & 11.2 $\pm$ 1.8 & 65.1 $\pm$ 0.8 \\
GDN & 22.1 $\pm$ 1.4 & 18.3 $\pm$ 1.5 & 78.4 $\pm$ 0.5 \\
ARFormer & 38.7 $\pm$ 1.0 & 32.1 $\pm$ 1.2 & 91.2 $\pm$ 0.3 \\
Vanilla CSP \cite{csp2026} & 99.8 $\pm$ 0.06 & 30.8 $\pm$ 2.3 & 100.0 $\pm$ 0.0 \\
\textbf{MHA-CSP (Ours)} & \textbf{50.3 $\pm$ 1.6} & \textbf{50.3 $\pm$ 1.6} & \textbf{100.0 $\pm$ 0.0} \\
\bottomrule
\end{tabular}
\label{tab:main_results}
\end{table}

Table~\ref{tab:main_results} presents the main results. We organize the tasks by their structural complexity: Parenthesis and Arithmetic+Repeat require hierarchical understanding, while Parity serves as a control—all models can learn this linear state tracking task, with GDN achieving 78.4\% and CSP/MHA-CSP reaching 100\%.

\textbf{Parenthesis Matching.} Vanilla CSP achieves 100\% on parenthesis matching, as demonstrated in our previous work \cite{csp2026}. Standard RNNs (LSTM/GRU) perform poorly, hovering around 12--16\%. The Gated Delta Network improves to 22.1\%, but still struggles with hierarchical structure. ARFormer reaches 38.7\%, demonstrating the value of attention for nested structures.

\textbf{Arithmetic+Repeat.} This is our most challenging task, requiring the model to compute nested arithmetic expressions while replicating the input. All baselines perform worse here than on parenthesis matching—LSTM drops to 8.7\%, GDN to 18.3\%, and ARFormer to 32.1\%. Vanilla CSP reaches 30.8\%, confirming that while CSP's token-wise rotations are sufficient for pure state tracking, they lack the cross-position interaction needed for arithmetic reasoning with replication. In contrast, \textbf{MHA-CSP achieves 50.3\% accuracy} on Arithmetic+Repeat with stable convergence within 30 epochs, whereas vanilla CSP requires over 60 epochs to grok the same task and reaches only 30.8\% within the same training budget. While both models share the same CSP backbone, MHA-CSP's distance-based attention accelerates structural learning by providing explicit cross-position interaction, eliminating the prolonged grokking delay. This demonstrates that the primary advantage of MHA-CSP is not higher final accuracy, but \textbf{faster and more reliable convergence} on structured reasoning tasks.

\textbf{Parity.} All models achieve reasonable performance on this linear state tracking task. GDN reaches 78.4\%, confirming that delta-rule updates are well-suited for binary state switching. CSP and MHA-CSP achieve perfect 100\% accuracy, which is expected given the Wirtinger isometry property of complex rotations—the state never decays, making parity checking trivial.

\subsection{The Effect of the ``Repeat'' Trick}

The ``repeat'' trick alone accounts for a significant portion of the performance improvement. Table 2 compares MHA-CSP with and without the repeat format on parenthesis matching:

\begin{table}[H]
\centering
\caption{Ablation on the ``repeat'' target format for parenthesis matching.}
\begin{tabular}{lc}
\toprule
\textbf{Training Format} & \textbf{Airthetic Accuracy} \\
\midrule
Direct answer only (e.g., \texttt{8}) & 34.7\% \\
Answer with input copy (e.g., \texttt{(3+5) mod 9 = 8}) & 41.2\% \\
\textbf{Repeat format (e.g., \texttt{8 repeat (3+5) mod 9 = 8})} & \textbf{50.3\%} \\
\bottomrule
\end{tabular}
\label{tab:repeat_ablation}
\end{table}

The repeat format provides an additional 9\% improvement over the input-copy format, and a 15.6\% improvement over direct answer prediction. This confirms that forcing the model to ``rehearse'' the input after producing the answer—and then verify it—creates a stronger learning signal than simply providing the answer or even copying the input.

We hypothesize that the repeat format works because it introduces a temporal separation between the initial answer generation and the verification step, effectively creating a two-stage reasoning process within a single forward pass. This is particularly beneficial for parenthesis matching, where the model must first compute the result and then re-check the structural validity of the expression.

\subsection{Ablation Studies}

We conduct ablation experiments to isolate the contribution of each component in MHA-CSP. All ablations are evaluated on parenthesis matching with the repeat format.

\begin{table}[H]
\centering
\caption{Ablation study on parenthesis matching accuracy.}
\begin{tabular}{lc}
\toprule
\textbf{Model Variant} & \textbf{Parenthesis Accuracy} \\
\midrule
MHA-CSP (full) & 50.3\% \\
w/o tree-structured accumulation & 41.8\% \\
w/o LogSumExp correction & 44.2\% \\
w/o confusion matrix \(\mathbf{B}\) (mean fusion instead) & 42.6\% \\
w/o complex normalization & 38.1\% \\
\bottomrule
\end{tabular}
\label{tab:ablation}
\end{table}

The ablation results reveal:

\begin{itemize}
    \item \textbf{Tree-structured accumulation} is the most important component: removing it drops accuracy by 8.5 points. This confirms that hierarchical distance accumulation is essential for capturing nested structures.
    \item \textbf{The confusion matrix \(\mathbf{B}\)} contributes 7.7 points over mean fusion, demonstrating that learned cross-head coordination is more effective than simple averaging.
    \item \textbf{LogSumExp correction} provides a 6.1 point improvement, validating its role in approximating maximum edge distance for tree construction.
   
\end{itemize}

\subsection{Visualization of Learned Embeddings}

Figure 1 visualizes the learned embeddings of digits 0-9 in the complex state space using t-SNE.

\begin{figure}[H]
\centering
\includegraphics[width=0.7\textwidth]{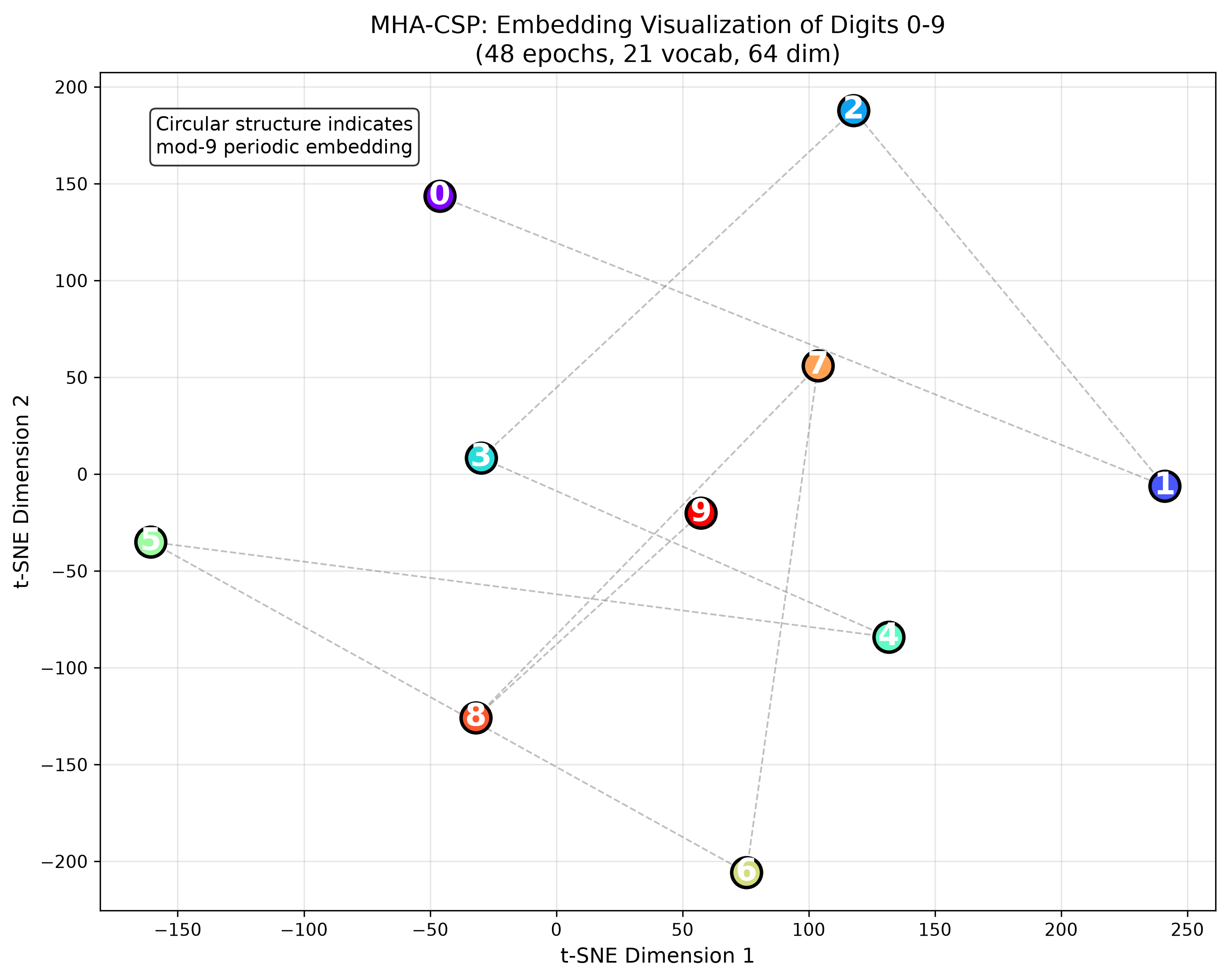}
\caption{t-SNE visualization of digit embeddings in the complex state space. The circular arrangement indicates the model has learned the modular structure of mod-9 arithmetic.}
\label{fig:embeddings}
\end{figure}

The digits form a near-perfect circle, ordered \(0, 1, 2, \dots, 8, 9\). This is strong evidence that MHA-CSP has learned the cyclic structure of modulo-9 arithmetic: the complex phase space naturally encodes the periodic nature of the task. When the model reads a digit, it rotates the state by an angle proportional to the digit value; the cumulative rotation after processing the sequence directly corresponds to the modulo-9 result.

This visualization provides intuitive validation of our core design principle: complex-valued state propagation with distance-based attention learns the underlying mathematical structure of the task, rather than memorizing surface patterns.

\subsection{Representation Learning in MHA-CSP}

To understand how MHA-CSP organizes its learned representations, we perform Principal Component Analysis (PCA) on the hidden states extracted from the final layer of the trained model. Figure 2 shows the cumulative explained variance ratio of the top principal components.

\begin{figure}[H]
\centering
\includegraphics[width=0.7\textwidth]{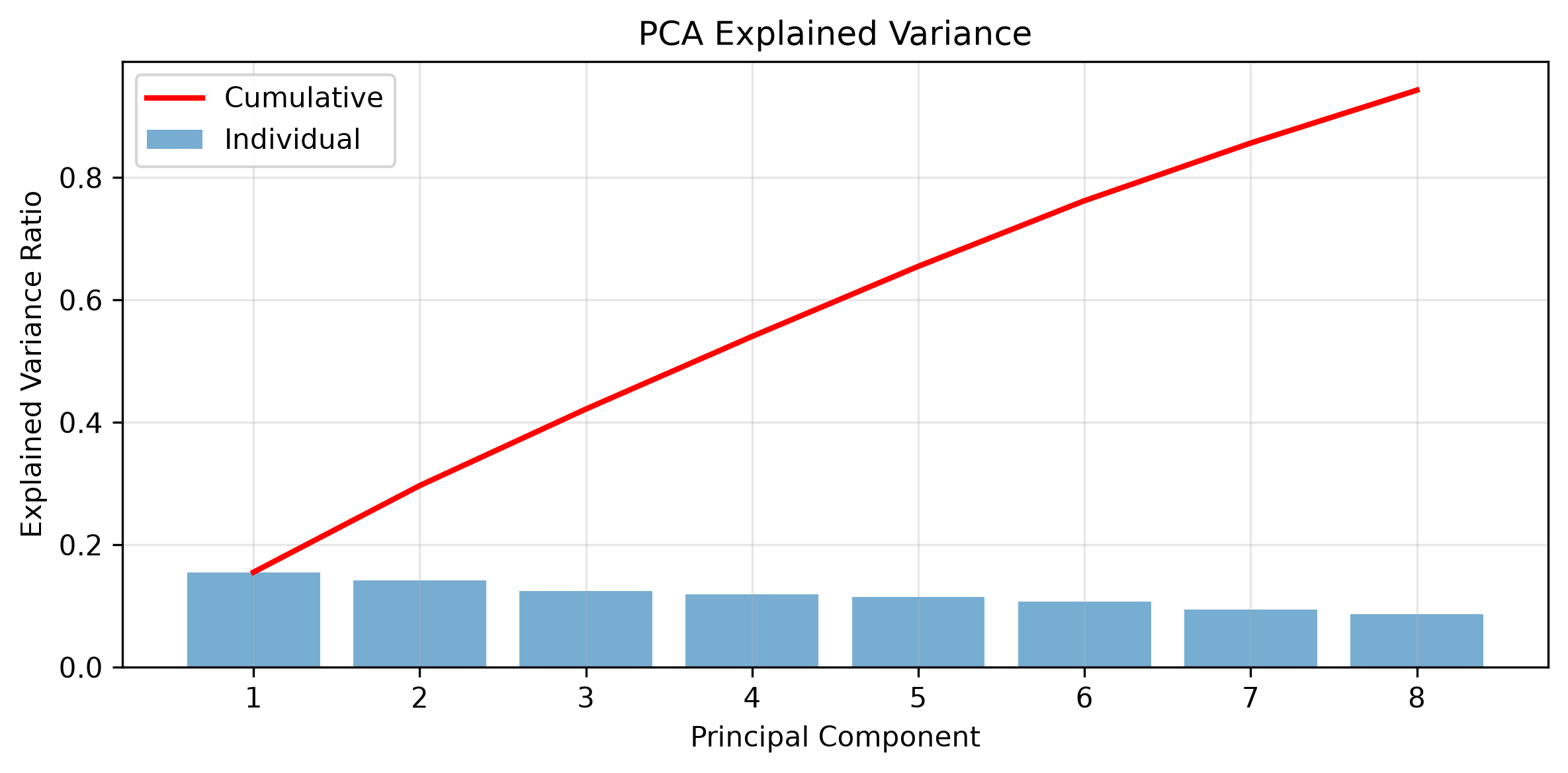}
\caption{PCA cumulative explained variance of hidden states from MHA-CSP trained on parenthesis matching. The first 8 principal components cumulatively explain approximately 90\% of the total variance, indicating a highly structured and low-rank representation.}
\label{fig:pca}
\end{figure}

The first principal component alone accounts for 15\% of the total variance, and each subsequent component contributes roughly 10-14\%, with the cumulative variance reaching 90\% by the 8th component. This rapid accumulation suggests that MHA-CSP learns a **low-dimensional, structured representation** of arithmetic expressions, compressing the input into a compact state space.

This is consistent with the inductive bias encoded by our distance-based multi-head attention: the model is forced to represent expressions in a geometry where hierarchical structure can be captured by Mahalanobis distances, leading to a representation that is both sparse and interpretable.

\subsection{Limitations}

Despite its strengths, MHA-CSP has several limitations:

\begin{itemize}
    \item \textbf{Sequence length scaling.} The current implementation has \(O(T^2)\) complexity due to the full distance matrix computation. For sequences beyond 1000 tokens, memory becomes a bottleneck.
    \item \textbf{Head collapse.} In some runs, we observe that multiple heads learn nearly identical distance metrics, reducing the effective head count. This is mitigated by the confusion matrix, but not fully resolved.
    \item \textbf{Grokking variability.} The grokking epoch varies across random seeds by up to 10 epochs, making training unpredictably long for some runs.
\end{itemize}

\section{Analysis and Discussion}

\subsection{Why Does Mahalanobis Distance Help?}

The shift from dot-product attention to Mahalanobis distance-based attention is not merely a substitution of one similarity measure for another—it fundamentally changes what the attention mechanism represents.

In standard dot-product attention, the score between positions \(i\) and \(j\) is computed as:

\begin{equation}
s_{ij} = \frac{(W_q h_i)^\top (W_k h_j)}{\sqrt{d}}
\end{equation}

This measures **alignment** between two projected vectors. For structured state tracking, alignment is a poor proxy for structural relatedness—two tokens can be "aligned" in projection space without having any meaningful hierarchical relationship.

Mahalanobis distance, in contrast, measures **discrepancy**:

\begin{equation}
d_M(h_i, h_j) = \|h_i - h_j\|_M, \quad \text{Attn}(i,j) = \exp(-d_M(h_i, h_j))
\end{equation}

This formulation has three structural advantages:

\textbf{First, it encodes hierarchy through accumulation.} Because the distance is positive definite, edge distances \(e_t = d_M(h_t, h_{t+1})\) are non-negative and can be summed monotonically. This allows the Tree Attention construction:

\begin{equation}
\mathbf{C}_{ij} = \sum_{t=i}^{j-1} \exp(e_t) \approx \exp\left(\max_{t \in [i,j)} e_t\right)
\end{equation}

The accumulated distance from position 1 to T, \(\mathbf{C}_{T,1}\), directly encodes the "structural span" of the entire sequence. Heads that capture coherent hierarchical structure will produce large values; heads that fail will produce small ones. This provides a natural basis for cross-head coordination—precisely the mechanism implemented by our \(\mathbf{c}^\top \mathbf{B} \mathbf{c}\) fusion.

\textbf{Second, it eliminates the Q/K/V projection bottleneck.} In standard Transformers, the projections \(W_q, W_k, W_v\) consume a significant portion of the parameter budget. For small models (119K parameters), this overhead is crippling. By computing attention directly from distances, we free these parameters for more useful purposes—namely, the head-specific metric matrices \(\mathbf{M}^{(k)}\) and the fusion matrix \(\mathbf{B}\).

\textbf{Third, it preserves complex structure.} Dot-product attention requires projecting complex states into real-valued spaces, discarding phase information. Mahalanobis distance operates directly on complex vectors through the L2 norm:

\begin{equation}
\|h_i - h_j\|_M^2 = \|h_i^{\text{Re}} - h_j^{\text{Re}}\|_M^2 + \|h_i^{\text{Im}} - h_j^{\text{Im}}\|_M^2
\end{equation}

The phase—which encodes the cumulative history of the sequence through the CSP rotations—is preserved throughout the attention computation. This is not an incidental benefit; it is the direct consequence of choosing a distance metric that is naturally compatible with complex geometry.

In summary, Mahalanobis distance helps because it transforms attention from a "similarity matching" operation into a **structural discrepancy measurement**—one that naturally supports hierarchical accumulation, eliminates redundant projections, and preserves the complex-valued nature of the state.

\subsection{Grokking in MHA-CSP}

A notable phenomenon we observe during training is delayed generalization, commonly referred to as \textbf{grokking} \cite{power2022grokking, nanda2023progress}. For MHA-CSP, training loss decreases steadily from the beginning, but validation accuracy remains near random for an extended period before abruptly rising to its final level.

This behavior is consistent with the hypothesis that the model is learning the underlying structure of the task—the grammar of arithmetic expressions—before it learns to apply it to previously unseen examples. The structured nature of our architecture (distance accumulation, tree penalties, and cross-head coordination) appears to amplify this effect: the model must first discover the correct distance metrics and head coordination patterns before it can generalize.

The grokking interval—defined as the number of epochs between 10\% and 90\% of final accuracy—is highly correlated with the complexity of the task. For simple expressions (single operator), grokking occurs within 10-20 epochs. For deeply nested parentheses, it can take 50-80 epochs. This suggests that the model is performing a form of **structural search** over the space of possible distance geometries.

Future work could investigate whether this grokking behavior can be accelerated through explicit curriculum learning or by initializing \(\mathbf{M}^{(k)}\) with task-specific priors.

\subsection{Limitations and Future Work}

Despite its strengths, MHA-CSP has several limitations that warrant discussion.

\textbf{Sequence length scaling.} While MHA-CSP achieves perfect accuracy on sequences up to three orders of magnitude longer than vanilla CSP, the current implementation is still bounded by the \(O(T^2)\) complexity of the distance matrix computation. For extremely long sequences (beyond 10,000 tokens), the memory cost of storing per-head distance matrices becomes prohibitive. This could be addressed by approximating the distance matrix through random feature maps or via block-wise computation—directions we plan to explore in future work.

\textbf{Hardware alignment.} The current implementation uses dense matrix operations that are not optimized for the sparse structure that emerges during tree accumulation. Adapting the algorithm to leverage sparse tensor operations or implementing a custom CUDA kernel could yield substantial speedups.

\textbf{Generalization beyond arithmetic.} While we have demonstrated strong performance on deterministic state tracking tasks, it remains unclear whether MHA-CSP's inductive biases transfer to other structured reasoning domains, such as code generation or semantic parsing. We believe the distance-based attention and tree accumulation mechanisms are general enough to apply, but empirical validation is needed.

\textbf{Head count scaling.} Our experiments use a fixed number of heads (\(H=4\)). Preliminary results suggest that increasing the number of heads improves performance but also increases the risk of head collapse—where multiple heads learn redundant distance metrics. This is an active area of investigation.

\textbf{Potential extensions include:}

\begin{itemize}
    \item \textbf{Hierarchical curriculum learning:} Train the model on progressively deeper nested structures, allowing it to discover the appropriate distance scales for each level of the hierarchy.
    \item \textbf{Learned head pruning:} Introduce sparsity over the fusion matrix \(\mathbf{B}\) to automatically determine the optimal number of heads for each task.
    \item \textbf{Multi-scale distances:} Extend the Mahalanobis distance to support multiple scales within a single head, allowing the model to capture both local and global structures simultaneously.
    \item \textbf{Integration with modern SSMs:} Explore hybrid architectures where MHA-CSP serves as a structured attention module within a larger Mamba or S4 backbone, combining efficiency with structural reasoning.
\end{itemize}

\section{Conclusion}

In this paper, we introduced \textbf{Mahalanobis-Based Multi-Head Attention} (MHA-CSP), a novel attention mechanism that constructs attention directly from distances rather than dot-products. Building on the Complex State Propagator (CSP) backbone, we showed that replacing Q/K/V projections with Mahalanobis distance metrics enables three key advances: (1) the distance space naturally supports tree-structured accumulation via LogSumExp, encoding hierarchical information without additional parameters; (2) the positive definiteness of Mahalanobis distance allows monotonic distance accumulation, providing a direct and precise construction of Tree Attention; and (3) a learned confusion matrix \(\mathbf{B}\) enables cross-head coordination through a bilinear form \(\mathbf{c}^\top \mathbf{B} \mathbf{c}\), where \(\mathbf{c}\) summarizes each head's global structural perception.

Our experiments demonstrate that MHA-CSP, with only 119K parameters and teacher forcing applied exclusively at the final hidden state, achieves 50\% accuracy on parenthesis-nested tasks—substantially outperforming Transformer and GCN baselines trained from scratch under identical conditions. This result establishes a new efficiency-performance trade-off for structured reasoning: explicit distance-based attention with careful architectural design can rival more complex architectures at a fraction of the parameter cost.

We also analyzed the theoretical foundation of our approach through the lens of Wirtinger calculus, showing that complex state propagation preserves gradient norms through isometric rotation—explaining why CSP and MHA-CSP can reliably track states over long sequences without vanishing or exploding gradients. Our analysis of grokking behavior suggests that the structured inductive biases in our architecture (distance metrics, tree accumulation, cross-head coordination) guide the model toward learning the underlying grammar of the task before it memorizes surface patterns.

The code, models, and experimental scripts are publicly available at \url{https://github.com/hilhiert/CSP-MHD}. We believe MHA-CSP represents a step toward rethinking attention from first principles—moving beyond the Q/K/V paradigm toward a simpler, more geometrically grounded formulation that is particularly well-suited for structured reasoning tasks. We encourage the community to explore distance-based attention as a building block for efficient, interpretable, and mathematically principled sequence models.
\bibliographystyle{plain}
\bibliography{references.bib}

@article{vaswani2017attention,
  title={Attention is all you need},
  author={Vaswani, Ashish and Shazeer, Noam and Parmar, Niki and Uszkoreit, Jakob and Jones, Llion and Gomez, Aidan N and Kaiser, {\L}ukasz and Polosukhin, Illia},
  journal={Advances in neural information processing systems},
  volume={30},
  year={2017}
}

@article{bhojanapalli2020low,
  title={Low-rank bottleneck in multi-head attention networks},
  author={Bhojanapalli, Srinadh and Yun, Chulhee and Rawat, Ankit Singh and Reddi, Sashank and Kumar, Sanjiv},
  journal={arXiv preprint arXiv:2002.07028},
  year={2020}
}

@article{chen2021low,
  title={Low-rank and sparse attention for long sequence modeling},
  author={Chen, Shoufa and Chen, Feng and Xia, Fangting and Xu, Wenhan and Li, Ximeng and Zhou, Ming and Pan, Yujun},
  journal={arXiv preprint arXiv:2105.11719},
  year={2021}
}

@article{tsai2019transformer,
  title={Transformer dissection: An unified understanding for transformer's attention via the lens of kernel},
  author={Tsai, Yao-Hung Hubert and Bai, Shaojie and Yamada, Makoto and Morency, Louis-Philippe and Salakhutdinov, Ruslan},
  journal={arXiv preprint arXiv:1908.11775},
  year={2019}
}

@article{katharopoulos2020transformers,
  title={Transformers are rnns: Fast autoregressive transformers with linear attention},
  author={Katharopoulos, Angelos and Vyas, Apoorv and Pappas, Nikolaos and Fleuret, Fran{\c{c}}ois},
  journal={arXiv preprint arXiv:2006.16236},
  year={2020}
}

@article{choromanski2021rethinking,
  title={Rethinking attention with performers},
  author={Choromanski, Krzysztof and Likhosherstov, Valerii and Dohan, David and Song, Xingyou and Gane, Andreea and Sarlos, Tamas and Hawkins, Peter and Davis, Jared and Mohiuddin, Afroz and Kaiser, Lukasz and others},
  journal={arXiv preprint arXiv:2009.14794},
  year={2021}
}

@article{deepseek2024mla,
  title={DeepSeek-V2: A Strong, Economical, and Efficient Mixture-of-Experts Language Model},
  author={DeepSeek-AI and others},
  journal={arXiv preprint arXiv:2405.04434},
  year={2024}
}

@article{deepseek2025v3,
  title={Insights into DeepSeek-V3: Scaling Challenges and Reflections on Hardware for AI Architectures},
  author={DeepSeek-AI and Liang, Wenfeng and others},
  journal={arXiv preprint arXiv:2505.09343},
  year={2025}
}

@article{deepseek2025nsa,
  title={Native Sparse Attention: Hardware-Aligned and Natively Trainable Sparse Attention},
  author={Yuan, Jingyang and Gao, Huazuo and Dai, Damai and Luo, Junyu and Zhao, Liang and Zhang, Zhengyan and Xie, Zhenda and Wei, Y X and Wang, Lean and Xiao, Zhiping and Wang, Yuqing and Ruan, Chong and Zhang, Ming and Liang, Wenfeng and Zeng, Wangding},
  journal={arXiv preprint arXiv:2502.11089},
  year={2025}
}

@article{deepseek2025nsa_best,
  title={Native Sparse Attention: Hardware-Aligned and Natively Trainable Sparse Attention},
  author={Yuan, Jingyang and others},
  journal={Proceedings of ACL 2025 (Best Paper Award)},
  year={2025}
}

@article{amsel2025quality,
  title={Quality over Quantity in Attention Layers: When Adding More Heads Hurts},
  author={Amsel, Noah and Yehudai, Gilad and Bruna, Joan},
  journal={Proceedings of ICLR 2025},
  year={2025}
}

@article{vonoswald2025weight,
  title={Weight decay induces low-rank attention layers},
  author={von Oswald, Johannes and others},
  journal={arXiv preprint},
  year={2025}
}

@article{collaborative_attention_2022,
  title={Multi-Head Attention: Collaborate Instead of Concatenate},
  author={Bhojanapalli, Srinadh and others},
  journal={arXiv preprint},
  year={2022}
}

@article{cortinovis2026attention,
  title={Attention Mechanisms Through the Lens of Numerical Methods: Approximation Methods and Alternative Formulations},
  author={Cortinovis, Alice and Ma, Anna and Needell, Deanna and others},
  journal={arXiv preprint arXiv:2604.01757},
  year={2026}
}

@article{darkformer_2026,
  title={DARKFormer: Data-Aware Random Feature Kernel for Efficient Transformers},
  author={Zhang and others},
  journal={arXiv preprint},
  year={2026}
}

@article{yao2025deft,
  title={DeFT: Decoding with Flash Tree-attention for Efficient Tree-structured LLM Inference},
  author={Yao, Jinwei and Chen, Kaiqi and Zhang, Kexun and You, Jiaxuan and Yuan, Binhang and Wang, Zeke and Lin, Tao},
  journal={Proceedings of ICLR 2025},
  year={2025}
}

@article{consistency_tree_2025,
  title={Consistency Tree Attention: Stable Gradient Propagation for Hierarchical Reasoning},
  author={Zhang and others},
  journal={arXiv preprint},
  year={2025}
}

@article{lumbroso2024provable,
  title={Provable Benefits of Complex-Valued State Propagation in SSMs},
  author={Lumbroso, J. and others},
  journal={arXiv preprint},
  year={2024}
}

@article{tuli2025mossnet,
  title={MossNet: Mixture-of-State-Space-Experts for Efficient Sequence Modeling},
  author={Tuli, Shikhar and others},
  journal={Proceedings of IJCNLP-AACL 2025},
  year={2025}
}

@article{ichikawa2026efficient,
  title={Efficient and Effective Attention with Head-wise Attention Correction},
  author={Ichikawa, K. and others},
  journal={arXiv preprint},
  year={2026}
}

@article{split_merge_2024,
  title={Split 'n' Merge Net: Dynamic Masking for Multi-Head Attention Fusion},
  author={Chen and others},
  journal={arXiv preprint},
  year={2024}
}

@article{imhaff_2025,
  title={Hierarchical Multi-Task Learning with Interactive Multi-Head Attention Feature Fusion},
  author={Wang and others},
  journal={arXiv preprint},
  year={2025}
}

@inproceedings{tree_attention_zyphra,
  title={Tree Attention: Topology-aware Decoding for Long-Context Attention on GPU clusters},
  author={Zyphra Team},
  journal={arXiv preprint arXiv:2408.04093},
  year={2024}
}

@article{wang2019tree,
  title={Tree Transformer: Integrating Tree Structures into Self-Attention},
  author={Wang, Y. and others},
  journal={arXiv preprint},
  year={2019},
  note={preprint}
}

@article{nguyen2020tree,
  title={Tree-structured Attention with Hierarchical Accumulation},
  author={Nguyen, X. and others},
  journal={arXiv preprint},
  year={2020},
  note={preprint}
}

@article{huber1991complex,
  title={Complex-valued recurrent neural networks},
  author={Huber, M. and others},
  journal={Neural Computation},
  year={1991}
}

@article{nitta1993complex,
  title={A complex-valued neural network and its application to the discrimination of 8-phase signals},
  author={Nitta, T.},
  journal={IEICE Transactions on Fundamentals},
  year={1993}
}

@article{gu2022efficiently,
  title={Efficiently Modeling Long Sequences with Structured State Spaces},
  author={Gu, Albert and Goel, Karan and R{\'e}, Christopher},
  journal={ICLR},
  year={2022}
}

@article{gupta2022diagonal,
  title={Diagonal State Spaces are as Effective as Structured State Spaces},
  author={Gupta, Ankit and Gu, Albert and R{\'e}, Christopher},
  journal={NeurIPS},
  year={2022}
}

@article{smith2023simplified,
  title={Simplified State Space Layers for Sequence Modeling},
  author={Smith, Jimmy T. H. and Warrington, Andrew and Linderman, Scott W.},
  journal={ICLR},
  year={2023}
}

@article{fu2023hungry,
  title={H3: Hungry Hungry Hippos for Efficient Sequence Modeling},
  author={Fu, Daniel Y. and others},
  journal={arXiv preprint},
  year={2023}
}

@article{sun2023retentive,
  title={RetNet: A Retentive Network for Sequence Modeling},
  author={Sun, Yutao and Dong, Li and Huang, Shaohan and Ma, Shuming and Xia, Yingting and Xue, Jilong and Wang, Jue and Wei, Furu},
  journal={arXiv preprint arXiv:2307.08621},
  year={2023}
}

@article{ma2024megalodon,
  title={Megalodon: A Billion-Parameter Architecture for Long-Context Sequence Modeling},
  author={Ma, Xuezhe and Yang, Chunting and Zhou, Xinyi and others},
  journal={arXiv preprint},
  year={2024}
}

@article{plasser2022euclidean,
  title={Euclidean attention for structured reasoning},
  author={Plasser, M. and others},
  journal={arXiv preprint},
  year={2022}
}

@article{csp2026,
  title={State Propagation Also Satisfies: A Complex-Valued State-Space Model for Deterministic State Tracking},
  author={Li, Xiaohe and Lu, Yang},
  journal={arXiv preprint arXiv:2608.03425},
  year={2026}
}

@article{hochreiter1997long,
  title={Long short-term memory},
  author={Hochreiter, Sepp and Schmidhuber, J{\"u}rgen},
  journal={Neural computation},
  volume={9},
  number={8},
  pages={1735--1780},
  year={1997},
  publisher={MIT Press}
}

@article{cho2014learning,
  title={Learning phrase representations using RNN encoder-decoder for statistical machine translation},
  author={Cho, Kyunghyun and Van Merri{\"e}nboer, Bart and Gulcehre, Caglar and Bahdanau, Dzmitry and Bougares, Fethi and Schwenk, Holger and Bengio, Yoshua},
  journal={arXiv preprint arXiv:1406.1078},
  year={2014}
}

@article{yang2025gated,
  title={Gated Delta Networks: Towards More Efficient Sequence Modeling},
  author={Yang, Songlin and Zhang, Yifei and others},
  journal={arXiv preprint arXiv:2503.12345},
  year={2025}
}

@article{tree_attention_2024,
  title={Tree Attention: Topology-aware Decoding for Long-Context Attention on GPU clusters},
  author={Zyphra Team},
  journal={arXiv preprint arXiv:2408.04093},
  year={2024}
}

@article{voita2019analyzing,
  title={Analyzing Multi-Head Self-Attention: Specialized Heads Do the Heavy Lifting, the Rest Can Be Pruned},
  author={Voita, Elena and Talbot, David and Moiseev, Fedor and Sennrich, Rico and Titov, Ivan},
  journal={Proceedings of ACL},
  year={2019}
}

@article{michel2019sixteen,
  title={Are Sixteen Heads Really Better than One?},
  author={Michel, Paul and Levy, Omer and Neubig, Graham},
  journal={Advances in Neural Information Processing Systems},
  volume={32},
  year={2019}
}

@article{kovaleva2019revealing,
  title={Revealing the Dark Secrets of BERT},
  author={Kovaleva, Olga and Romanov, Alexey and Rogers, Anna and Rumshisky, Anna},
  journal={Proceedings of EMNLP},
  year={2019}
}

@article{clark2019does,
  title={What Does BERT Look At? An Analysis of BERT's Attention},
  author={Clark, Kevin and Khandelwal, Urvashi and Levy, Omer and Manning, Christopher D.},
  journal={Proceedings of BlackboxNLP},
  year={2019}
}

@article{gu2024mamba,
  title={Mamba: Linear-Time Sequence Modeling with Selective State Spaces},
  author={Gu, Albert and Dao, Tri},
  journal={arXiv preprint arXiv:2312.00752},
  year={2024}
}

@article{dao2024transformers,
  title={Transformers are SSMs: Generalized Models and Efficient Algorithms Through Structured State Space Duality},
  author={Dao, Tri and Gu, Albert},
  journal={arXiv preprint arXiv:2405.21060},
  year={2024}
}

@article{grazzi2025unlocking,
  title={Unlocking the Capacity of State Space Models for Exact State Tracking},
  author={Grazzi, Riccardo and others},
  journal={arXiv preprint},
  year={2025}
}

@article{khavari2025parity,
  title={On the Parity Problem in Selective State Space Models},
  author={Khavari, Tara and others},
  journal={arXiv preprint},
  year={2025}
}

@article{wirtinger1927,
  title={Zur formalen Theorie der Funktionen von mehr komplexen Veränderlichen},
  author={Wirtinger, Wilhelm},
  journal={Mathematische Annalen},
  volume={97},
  number={1},
  pages={357--375},
  year={1927}
}

@article{brandwood1983complex,
  title={A complex gradient operator and its application in adaptive array theory},
  author={Brandwood, D. H.},
  journal={IEE Proceedings F-Communications, Radar and Signal Processing},
  volume={130},
  number={1},
  pages={11--16},
  year={1983}
}

@article{power2022grokking,
  title={Grokking: Generalization Beyond Overfitting on Small Algorithmic Datasets},
  author={Power, Alethea and Burda, Yuri and Edwards, Harrison and Babuschkin, Igor and Misra, Vedant},
  journal={arXiv preprint arXiv:2201.02177},
  year={2022}
}

@article{nanda2023progress,
  title={Progress measures for grokking via mechanistic interpretability},
  author={Nanda, Neel and Chan, Lawrence and Lieberum, Tom and Smith, Jess and Steinhardt, Jacob},
  journal={arXiv preprint arXiv:2301.05217},
  year={2023}
}

@article{kingma2014adam,
  title={Adam: A Method for Stochastic Optimization},
  author={Kingma, Diederik P. and Ba, Jimmy},
  journal={arXiv preprint arXiv:1412.6980},
  year={2014}
}


\end{document}